\documentclass[sigconf]{acmart}
\usepackage{balance}

\AtBeginDocument{%
  \providecommand\BibTeX{{%
    \normalfont B\kern-0.5em{\scshape i\kern-0.25em b}\kern-0.8em\TeX}}}

\copyrightyear{2021} 
\acmYear{2021} 
\setcopyright{acmcopyright}\acmConference[MM '21]{Proceedings of the 29th ACM International Conference on Multimedia}{October 20--24, 2021}{Virtual Event, China}
\acmBooktitle{Proceedings of the 29th ACM International Conference on Multimedia (MM '21), October 20--24, 2021, Virtual Event, China}
\acmPrice{15.00}
\acmDOI{10.1145/3474085.3475237}
\acmISBN{978-1-4503-8651-7/21/10}

\begin{document}
\fancyhead{}
\title{Motion Prediction via Joint Dependency Modeling in Phase Space}

\author{Pengxiang Su}
\affiliation{%
  \institution{Jilin University}
  \city{Changchun}
  \state{Jilin}
  \country{China}
}
\email{supx19@mails.jlu.edu.cn}

\author{Zhenguang Liu}
\affiliation{%
  \institution{Zhejiang University}
  \city{Hangzhou}
  \state{Zhejiang}
  \country{China}}
\email{liuzhenguang2008@gmail.com}

\author{Shuang Wu}
\affiliation{%
  \institution{Nanyang Technological University}
  \country{Singapore}
}
\email{wushuang@outlook.sg}

\author{Lei Zhu}
\affiliation{%
 \institution{Shandong Normal Unversity}
 \city{Jinan}
 \state{Shandong}
 \country{China}}

\author{Yifang Yin}
\affiliation{%
  \institution{National University of Singapore}
  \country{Singapore}}

\author{Xuanjing Shen}
\affiliation{%
  \institution{Jilin University}
  \city{Changchun}
  \state{Jilin}
  \country{China}}
\email{xjshen@jlu.edu.cn}

\renewcommand{\shortauthors}{P.X. Su and Z.G. Liu, et al.}

\thanks{Corresponding Authors: Zhenguang Liu, Shuang Wu, Xuanjing Shen}

\begin{abstract}
Motion prediction is a classic problem in computer vision, which aims at forecasting future motion given the observed pose sequence. Various deep learning models have been proposed, achieving state-of-the-art performance on motion prediction. However, existing methods typically focus on modeling temporal dynamics in the pose space. Unfortunately, the complicated and high dimensionality nature of human motion brings inherent challenges for dynamic context capturing. Therefore, we move away from the conventional pose based representation and present a novel approach employing a phase space trajectory representation of individual joints. Moreover, current methods tend to only consider the dependencies between physically connected joints. In this paper, we introduce a novel convolutional neural model to effectively leverage \emph{explicit} prior knowledge of motion anatomy, and simultaneously capture both spatial and temporal information of joint trajectory dynamics. We then propose a global optimization module that learns the \emph{implicit} relationships between individual joint features.  

Empirically, our method is evaluated on large-scale 3D human motion benchmark datasets (\emph{i.e.}, Human3.6M, CMU MoCap). These results demonstrate that our method sets the new state-of-the-art on the benchmark datasets. Our code will be available at \url{https://github.com/Pose-Group/TEID}.
\end{abstract}

\ccsdesc[500]{Computing methodologies~Computer vision}
\ccsdesc[500]{Computing methodologies~Activity recognition and understanding}

\keywords{Motion prediction, trajectory, joint correlation.}


\maketitle

\section{Introduction}

Humans can effortlessly predict the future motions of animals or other humans. Such a crucial ability enables humans to intelligently interact with the external world~\cite{DMGNN,Long_Term_Prediction,Diverse_Motion_Prediction,DLow}. Similarly, in the domain of artificial intelligence, understanding and predicting future human motion is important, which enjoys a wide range of applications including human-robot interaction, intelligent driving, pose tracking, and motion generation~\cite{MoPredNet,BiHMP-GAN,ZhiyongCheng01,Adversarial_Geometry_Aware}.

\begin{figure}[t]
\begin{center}
\includegraphics[width=1\linewidth]{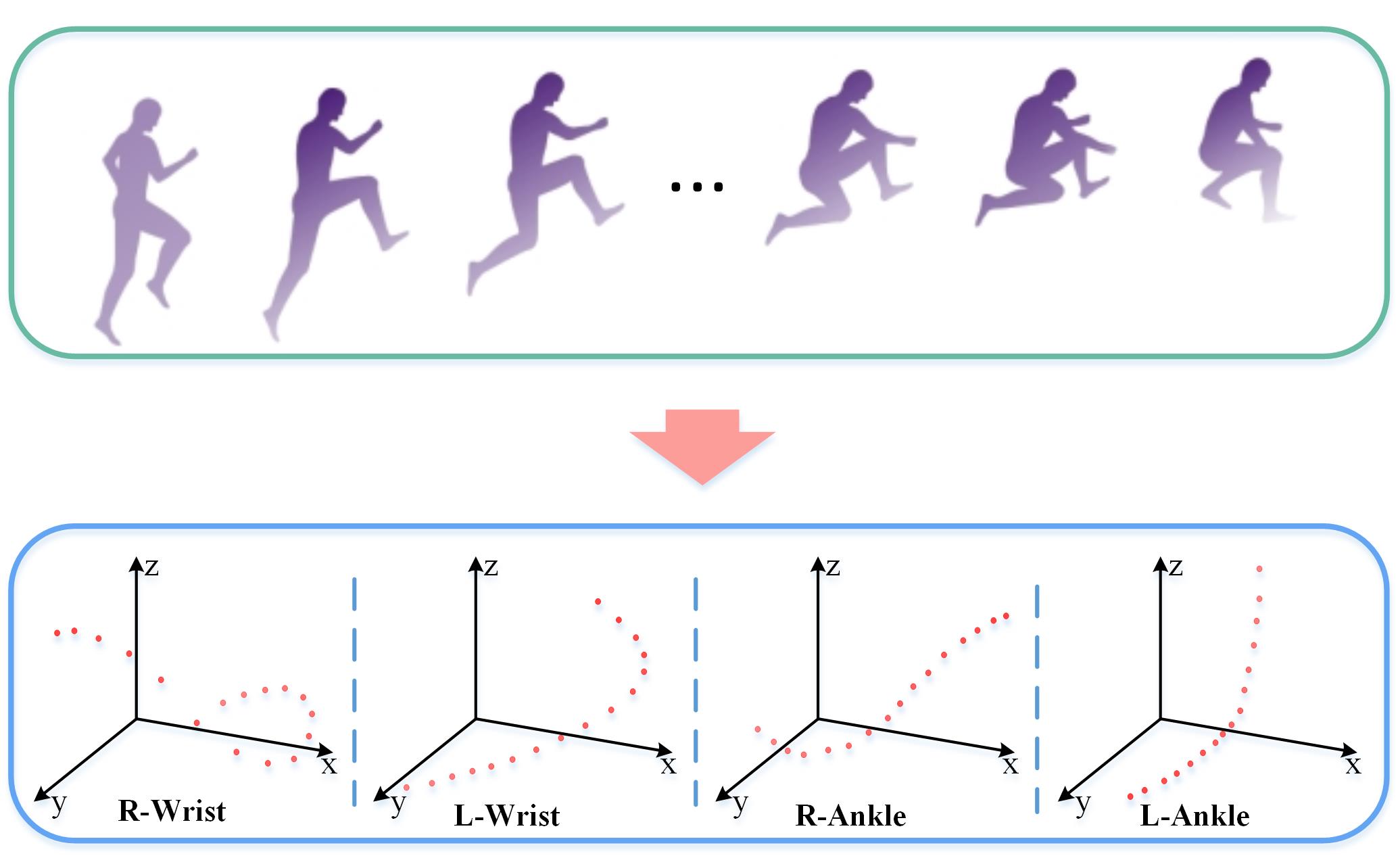}
\end{center}
   \caption{Trajectories of the wrist and ankle joints during jumping.}
\label{fig_jointtrajectory}
\end{figure}

Given the observed past motion data, which is typically represented as the 3D skeletal pose sequence, the \emph{human motion prediction} task aims to accurately estimate the future motion. 
Traditional approaches broadly fall within the scope of latent variable models, including Gaussian processes~\cite{Gaussian} and Markov models~\cite{Markov}. With the availability of large-scale labeled motion capture datasets, a number of neural network designs have been proposed, yielding much improved results. The general framework encompasses a sequence-to-sequence model whereby the observed sequence is encoded to a \emph{latent motion context} that is decoded to output future sequence.  One line of work \cite{erd,residual} utilized recurrent neural networks (RNNs) 
such as Long Short Term Memory (LSTM) to model motion contexts. Another
line of work \cite{LDR,DMGNN,FC-GCN} built upon the success of graph convolutional networks 
to better characterize the spatial connections between joints. Other works directed efforts at convolutional networks \cite{Conseq2seq}, hierarchical motion context representation \cite{cvprliu}, and attention-based networks \cite{HRI}, respectively.

Empirically, we implemented existing methods following their released code and scrutinized their results on benchmark datasets. Unfortunately, they still suffer from inaccurate motion predictions. We conjecture that the reasons are twofolds. 

\textbf{First, pose representation.} \quad Most existing works represent the historical motion sequence as a time series of the skeletal pose, parameterized as axis angles. A pose technically translates to the positions of all joints. Modeling motion contexts in the
pose space implicitly entails all the joints, putting motion prediction task on an unnecessarily high-dimensional manifold. On the other hand, we visualized the trajectories of each joint during motions, and observed that the trajectory of a joint is usually smooth. An illustrated example for the jumping motion is presented in Fig.~\ref{fig_jointtrajectory}, which confirms the smoothness of joint motions.
Furthermore, the results of \cite{FC-GCN} and our previous research \cite{Our} show that modeling motion contexts in the trajectory space can effectively promote the prediction accuracy.
Motivated by this, we present a new solution that works on the individual joint trajectory space, explicitly leveraging the smooth movement of a joint to predict its future. Specifically, we include the position of a joint and its \emph{displacements} (motion), \emph{i.e.}, the first-order differences of joint positions between adjacent frames. In doing so, we effectively represent the motion sequence in trajectory phase space, where both the current joint location configuration and joint frame-wise displacements are explicitly modeled. This enables a complete characterization of the dynamic system and the motion prediction task converts to trajectory extrapolation for each joint. Specifically, this representation focus on each joint rather than the entire pose, which characterizes the trajectory of a joint using frame-wise joint positions and augments the joint trajectory with instantaneous displacements. In this way, we explicitly encode motion semantics and the difficulty of learning the temporal evolution in the entire pose sequence is significantly reduced at the base level.

\textbf{Second, joint correlation modeling.} \quad On another front,  
the existence of laws of intersegmental dependencies is valuable for precise motion prediction. 
Although recent works have incorporated some form of inter-joint dependencies modeling through kinematic trees or graphs, there is insufficient tapping into the pose and motion prior. This deficiency manifests as inaccurate  portrayals of the subtleties 
in predicting different motions. We therefore design a novel spatio-temporal convolutional neural network to better encode prior knowledge, \emph{e.g., skeletal connections and motion semantics}. Specifically, the model is composed  of two phases. (1) In the first phase, we obtain a principled trajectory prediction of a joint by considering its own trajectory and trajectories of its explicitly correlated joints, which smoothes out the interferences of  irrelevant joints and saves computational burdens. Three parallel  convolution branches with different dilation rates are leveraged to extract multi-scale context  information.  (2) Since the trajectory extrapolation for each joint is handled individually so far, in the second phase we further engage in a global optimization module. The module  captures latent dependencies of the predicted future trajectory of each joint with respect to each other as well as the global skeletal motion. The individual joint trajectory is refined, thus ensuring consistency and a harmonious alignment of each joint within the whole.

\textbf{Contributions.} To summarize, the key contributions of this paper are as follows: 

\emph{(i)} We propose to learn motion dynamics in a new trajectory space and present a spatio-temporal convolutional feed-forward network to encode trajectory information for motion prediction.

\emph{(ii)} We introduce a two-phase model that  extracts the \emph{explicit} and \emph{implicit} relationships between joints.

\emph{(iii)} On two human motion benchmark datasets, including the large \emph{Human3.6M} and \emph{CMU MoCap} datasets, our method achieves the state-of-the-art performance in both short-term and long-term predictions.

The rest of the paper is organized as follows. We first review related work in Section 2.
We then introduce the details of our approach in Section 3. 
Thereafter, we evaluate the performance of the proposed method and compare it with existing methods 
in Section 4. Finally, we conclude the paper with summary and discussions on
the approach in Section 5.

\section{Related Work}
\subsection{Motion Prediction}

Previously, statistical approaches broadly fall within the scope of latent variable models, which leverage Gaussian processes~\cite{Gaussian} and hidden Markov models~\cite{Markov} to capture the temporal dynamic of human motions.  Recently, the success of deep learning methods in various fields \cite{ZhuLCLZ20,feng2019graph,yang2018person,dong2021dual,LiuSNK17,MMGCN,liu2021toward,LiuLWCHJ21} lead to diverse neural network designs for modeling motion contexts~\cite{ananLiu02,QianruSun01,zangao03,Dynamic_Future_Net}.
The general framework encompasses a sequence-to-sequence model, where the observed pose sequence is encoded to a latent motion context that is decoded to output future sequence. A major line of work utilizes RNN modules such as Long Short Term Memories (LSTM) \cite{erd} or Gated Recurrent Units (GRU) \cite{residual} as the encoder and decoder. Other options such as convolutional networks (CNN), hierarchical representation, and graph convolutional networks have also been explored. ~\cite{Conseq2seq} proposes a CNN based model to capture long-term temporal and spatial correlations.~\cite{cvprliu} represents human skeleton with a Lie algebra representation to encode anatomical constraints, and employs a hierarchical recurrent network to encode local and global contexts.~\cite{FC-GCN} designs a generic graph to encode the spatial dependencies of human pose and a simple feed-forward deep network to capture dynamic context.~\cite{LDR} builds a deep generative model based on a novel dynamic GNN and adversarial learning.
~\cite{DMGNN} introduces a multiscale graph computational unit to extract and fuse features for motion feature learning.~\cite{HRI} develops an attention-based model that explicitly leverages historical information for motion prediction.
In this paper, we present an alternative network, which improves the modeling of spatio-temporal contexts along with the capability of capturing both explicit and implicit inter-joint dependencies. 

\subsection{Skeleton-based Pose Representation}
A fundamental component for the motion prediction task is to represent the pose sequence in the most suitable and effective way so as to facilitate modeling of motion contexts. Most existing works \cite{erd,residual,Conseq2seq,cvprliu} represent the pose as joints along a kinematic graph with joint orientations parameterized as axis angles. A related approach characterizes the skeletal structure as a graph \cite{DMGNN,LDR,FC-GCN,HRI,Local_Structure_Representations}. A critical issue is that these schemes usually treat the joints on an equal standing, failing to account for the fact that the kinematic chain is a hierarchical structure. This raises severe difficulties in effectively capturing the joint dependencies, manifesting as large prediction errors for end effectors such as hands and foot. 
We therefore propose to discard with kinematic graphs, treat each joint as distinct entities, and directly learn the correlations and dependencies via data, facilitated by additional knowledge of skeletal anatomy.

\begin{figure*}[t]
\begin{center}
\includegraphics[width=1\textwidth]{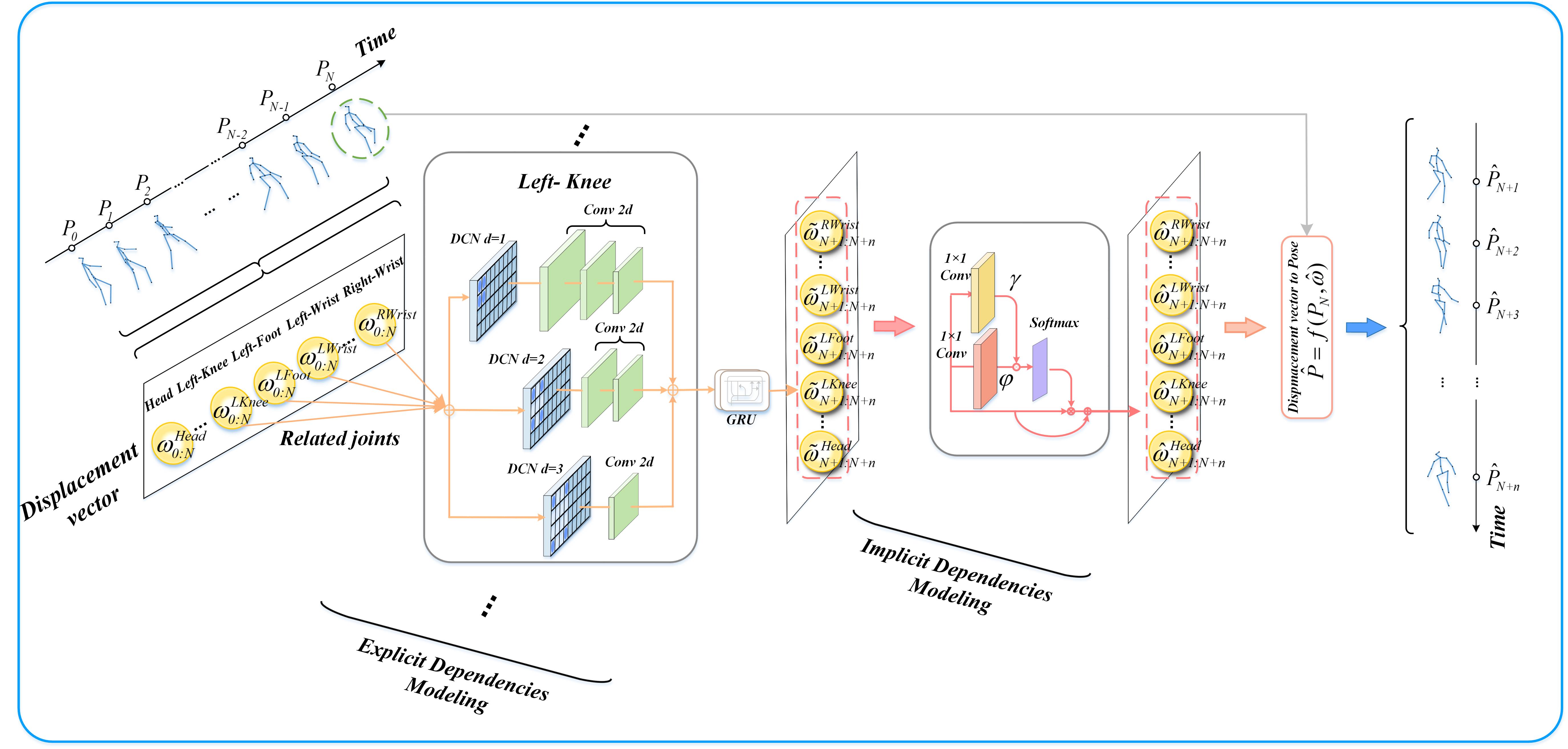}
\end{center}
   \caption{The proposed TEID network performs trajectory extrapolation in two phases. The first explicit dependency modeling phase consists of three deformable convolution branches with different dilation rates to incorporate explicit prior knowledge of joint relationships and extracts spatio features from joint phase space trajectories. A GRU network outputs intermediate displacement features which are then refined via our implicit dependency modeling phase, which captures the joint trajectory dependencies with respect to the global motion dynamics to improve holistic consistency. Finally, the future motion sequence is reconstructed from the extrapolated displacement features and the last observed pose.}
\label{architecture}
\end{figure*}

\section{The Proposed Method}

In this section, we provide the details of our approach, TEID (motion prediction with  phase space Trajectory representation and Explicit/Implicit Dependencies modeling). TEID can be divided into three key components. \emph{(i)} A historical pose sequence is converted into a series of joint displacement vectors in trajectory space. \emph{(ii)} These features are fed into a novel multi-scale convolutional network based on prior knowledge to capture the spatial and temporal dynamic context. \emph{(iii)} Finally, the predicted sequences are further input into a refinement network to improve the coordination of the entire sequence.

\subsection{Phase Space Trajectory Representation}
First, we move away from the widely adopted kinematic graph scheme and instead focus on each joint rather than the entire pose. In particular, a joint $j$ is characterized by its trajectory $T_j=[D^j_0,D^j_1,\cdots,D^j_N]$ where $D^j_i \in\mathbb{R}^3$ denotes the position vector of joint $j$ in the $i^{th}$ frame (time).
Furthermore, we augment a joint trajectory with joint instantaneous displacements by computing the joint displacements between adjacent frames, thereby obtaining phase space trajectory for each joint. Specifically, joint $j$ at frame $i$ is represented as a tuple
$(D^j_i,\omega^j_i)$ where $D^j_i$ characterizes its position while $\omega^j_i\in\mathbb{R}^3$  captures the displacement at the $i^{th}$ frame. Displacement $\omega^j_i$ can be conveniently computed by:
\begin{equation}
\begin{aligned}
\omega^j_i=D^j_{i}-D^j_{i-1}.
\end{aligned}
\end{equation}
Mathematically, the phase space is the cotangent bundle of the pose configuration space and by explicitly including the displacement we achieve a complete characterization of the dynamics of the system at any given frame (time) \cite{craig2009introduction}. The displacement information also provides valuable motion contexts for future prediction. \cite{residual} models angular velocities to solve the problem of the discontinuities between the observed sequence and the first-frame prediction. \cite{FC-GCN} tries to encode temporal information in the frequency space via the discrete cosine transform. 
Our proposed method instead uses the direct joint instantaneous displacements to explicitly model trajectory features, therefore, the network does not have to learn to extract the implicit motion dynamics of the joints.



\subsection{The Proposed Network}
An overview of the proposed TEID network is shown in Fig.~\ref{architecture}. An important highlight of our approach consists in rejecting the widely adopted kinematic graph based pose representation which inherently invokes an unnecessary complexity burden due to the hierarchical nature of the joints representation. Instead, we examine the phase space trajectories at individual joint levels. 
Taking phase space trajectory representations as inputs, TEID comprises an \emph{explicit modeling component} as well as an \emph{implicit modeling module} to capture \emph{explicit} and \emph{implicit} dependencies
between joint trajectories, respectively. This allows extrapolation and connection of the individual joint trajectories to form a single coherent pose sequence.

More specifically, TEID \textbf{first} models the known and explict dependencies between joints. Following \cite{Our}, three kinds of explicit joint dependencies are encoded: 1) The natural skeletal connection between
joints. 2) The correlation between symmetric arms and legs, \emph{e.g., the left arm and the right arm}. 3) The synchronization tendency between an arm and a leg in opposite sides, \emph{e.g., the left arm and the right leg}. \textbf{Then}, TEID considers the hidden and implicit denpendencies between joints and the naturalness of the entire pose.
The implicit dependency modeling encompasses a global optimization module that examines the implicit relationship between the predicted joint trajectory extrapolations and the entire predicted motion sequence to ensure consistency and  naturalness of the final output.

\textbf{Explicit Dependency Modeling} \quad 
Effectively capturing the precise spatial relationships between closely related  joints is indispensable to understanding and modeling motion. Existing methods have expended significant efforts in this direction such as considering kinematic trees \cite{cvprliu} or graphs \cite{DMGNN,LDR,HRI} and examining the problem in the frequency domain via discrete Fourier transforms \cite{FC-GCN}. Yet, the fixed graph or kinematic tree structures in such methods make it difficult to incorporate prior knowledge, while using frequency domain usually involves some intricate or cumbersome network designs to extract joint relationships. 

In our approach, prior anatomical knowledge of motion can be readily leveraged. Technically, explicit dependencies that can be tapped into include \emph{bone connections}, \emph{symmetrical properties of the arms and legs}, as well as \emph{more complicated examples such as coordination of different limbs in opposite sides}. As depicted in Fig.~\ref{architecture}, the phase trajectory inputs for a joint $j$ along with all its explicitly  related joints are collectively fed into a spatio-temporal module. 
Specifically, this convolutional module consists of three parallel layers of deformable convolutions with different dilation rates, which allow simultaneous extraction of multi-scale spatial features from the phase space trajectories. Various dilation rates $d \in \{1, 2, 3\}$ are engaged to facilitate multi-scale processing of the dynamic information. For each joint, the consideration of only its explicitly (closely) related joints rather than all the joints is beneficial in smoothing out noises and irrelevant interferences. Finally, we obtain for joint $j$ a feature of size $m\times C(j)\times N$ where $m$ is the number of channels for the convolution operation, $C(j)$ is the number of explicitly related joints pertaining to joint $j$, and $N+1$ is the length of historical sequence (which yields $N$ displacements).

The obtained features are then fed into a GRU network to generate principled trajectory predictions for future frames $N+1:N+n$.

\textbf{Implicit Dependency Modeling}  \quad  We further refine the principled trajectory prediction for each joint by engaging in \emph{implicit dependency modeling}, which considers the relationships between predicted trajectories of all joints. 
After the previous step, the predictions of different joint trajectories are independent to each other, the dependencies between a single joint and the entire pose sequence are also ignored. In order to ensure that the predicted motion sequence is harmonious and reasonable from a global perspective, we introduce a global refinement module which can optimize the predicted joint features with respect to the entire predicted pose sequence. 

As illustrated in Fig.~\ref{architecture}, we feed the forecasted displacement vector sequences $\tilde{\omega} \in \mathbb{R}{^{3 \times n \times J}}$ into two $1\times1$ convolution modules for generating two feature matrices $\gamma$ and $\varphi$, respectively. Then, we reshape them to $\mathbb{R}{^{3 \times K}}$, where $K=n\times J$ is the total number of joint positions in $n$ future frames. Formally,

\begin{equation}
\begin{aligned}
\tilde{\omega} \xrightarrow[\text{convolution}]{1\times1} \xrightarrow[]{reshape}\gamma\\
\tilde{\omega} \xrightarrow[\text{convolution}]{1\times1} \xrightarrow[]{reshape} \varphi.
\end{aligned}    
\end{equation}

Next, we apply matrix multiplication on the transpose of $\gamma$ and $\varphi$, and utilize a softmax operation to obtain a global affinity matrix $A\in \mathbb{R}^{K\times K}$:

\begin{equation}
\begin{aligned}
A = \frac{\exp(\gamma^{T} \cdot \varphi) }{\displaystyle\sum_{q = 1}^{K} \exp(\gamma^{T} \cdot \varphi)}.
\end{aligned}    
\end{equation}

The affinity matrix $A$ models the correlation between any two joints in the predicted $n$ future frames. Within the affinity matrix, element $A^{j,j'}_{i,i'}$ measures the influence of $j'^{th}$ joint in the $i'^{th}$ future frame on the  $j^{th}$ joint in the $i^{th}$ future frame. More precisely, the element of the matrix measures the affinity between any two joint displacement features even when the two joints are in different frames. Finally, we refine the predicted displacement vector feature of $j^{th}$ joint in the $i^{th}$ future frame (generated after explicit dependency modeling) by:


\begin{equation}\label{eq_finalw}
\begin{aligned}
\hat{\omega}^j_i = \tilde{\omega}^j_i + \sum_{j'=1, i'=N+1}^{J, N+n} A^{j,j'}_{i,i'} \tilde{\omega}^{j'}_{i'}.
\end{aligned}    
\end{equation}
where $\tilde{\omega}^j_i$ is the displacement prediction of the ${j}^{th}$ joint in the ${i}^{th}$ future frame, and $\tilde{\omega}^{j'}_{i'}$ is the displacement prediction of the ${j'}^{th}$ joint in ${i'}^{th}$ future frame, which might have influence on $\tilde{\omega}^j_i$. The final displacement vector $\hat{\omega}$ obtained by Eq.~\ref{eq_finalw} can be used to restore the new joint position by incorporating the joint position in the last observed frame. 




\subsection{Loss Function}
Previous works such as~\cite{LDR,HRI} have generally employed the standard L2 loss. However, the motion ranges of different joints are quite different. Some joints may stay motionless in an action while some joints may undergo large displacements. It is grossly inappropriate to assign the same weight to all the joints. We assign a spatio-temporal weight to each joint. We give higher weights to joints with larger historical position changes (which enforces the network to consider these prominent joints) and to earlier frames in the prediction (which is to reduce accumulation errors). 
Empirically, this is beneficial to improve the performance of short-term forecasting and can effectively reduce the error accumulation in long-term forecasting. Formally, the loss function can be formulated as:
\begin{equation}\label{loss}
\ell_{Traj}  = \sum_{i = N + 1}^{N + n} {\sum_{j = 1}^J {w^j_i} } \parallel \hat{\omega}^j_i - \omega^j_i{\parallel ^2}.
\end{equation}
where $\hat{\omega}^j_i$ is the displacement vector prediction of the ${j}^{th}$ joint in the ${i}^{th}$ frame, while $\omega^j_i$ represents the corresponding ground truth. $w^j_i$ denotes the weight. The loss enforces the predicted frame-wise displacements to approach the ground truth.

\section{Experiments}
In this section, we evaluate our method over the widely-used human motion benchmark datasets Human3.6M and CMU MoCap. 

\subsection{Datasets}
{\bf Human3.6M} is the largest dataset for human motion prediction task, containing 3.6 million human poses and corresponding videos (images). The videos are recorded by a vicon motion capture system. The dataset consists of 7 subjects, each of them performed 15 activities (\emph{e.g.}, discussion, eating, and sitting). There are 32 skeletal joints involved to characterize the human pose. Following the evaluation protocol of previous works \cite{erd}, we removed duplicate points, performed a down sampling to 25 frames per second (FPS), and set subject 5 as the test set, with training done on the remaining 6 subjects.

{\bf CMU MoCap} is released by Carnegie Mellon University. 12 infrared cameras record the position of 41 markers taped on the human body to capture 3D skeleton motion. We adopted the same preprocessing schema as that in \cite{FC-GCN}, 7 actions were selected for evaluating the performance of the model (\emph{e.g.}, walking, soccer, and jumping). 


\subsection{Experimental Settings}
We use PyTorch to implemente our method. For training, we utilize the ADAM optimizer with a batch size of 16, a learning rate of 0.001, and a dropout of 0.05. The model is trained for 50 epochs on a Nvidia GeForce Titan Xp GPU. To avoid the problem of exploding gradients, gradient clipping is utilized at a threshold of 5. We use a single convolutional feature channel for the explicit dependency modeling module and the GRU unit size is set to 128. 


\begin{table*}[t] 
\caption{Comparisons of position error for short-term and long-term predictions on H3.6m dataset. Our method consistently outperformance other methods.}
  \centering
  \resizebox{0.97\textwidth}{!}{
    \begin{tabular}{c|ccccccc|ccccccc|ccccccc}
    \hline
        \multicolumn{2}{r}{} &       & \multicolumn{4}{c}{Directions} & \multicolumn{2}{c}{} &       & \multicolumn{4}{c}{Eating} & \multicolumn{2}{c}{} &       & \multicolumn{4}{c}{Greeting} &
           \\
    Millisecond(ms) & 80    & 160   & 320   & 400    & 560    & 720   & 1,000  & 80    & 160   & 320   & 400  & 560    & 720   & 1,000 & 80    & 160   & 320   & 400  & 560    & 720   & 1,000  \\
    \midrule
    Res-GRU \cite{residual}          & 21.6  & 41.3  & 72.1  & 84.1  & 101.1          & 114.5            & 129.1  & 16.8  & 31.5  & 53.5  & 61.7  & 74.9         & 85.9        & 98.0   & 31.2   & 58.4  & 96.3  & 108.8  & 126.1       & 138.8       & 153.9\\
    ConSeq2Seq \cite{Conseq2seq}     & 13.5  & 29.0  & 57.6  & 69.7  & 86.6           & 99.8             & 115.8  & 11.0  & 22.4  & 40.7  & 48.4  & 61.3         & 72.8        & 87.1   & 22.0  & 45.0  & 82.0  & 96.0    & 116.9       & 130.7       & 147.3 \\
    HMR \cite{cvprliu}               & 23.3  & 25.0  & 47.2  & 61.5  & 80.9           & 95.1             & 116.9  & 9.2   & 13.9  & 34.6  & 47.1  & 61.3         & 72.9        & 84.8   & 12.9  & 31.9  & 55.6  & 82.5    & 104.3       & 116.1       & 123.2 \\
    FC-GCN \cite{FC-GCN}             & 12.6  & 24.4  & 48.2  & 58.4  & 72.2           & 86.7             & 105.8  & 8.8   & 18.9  & 39.4  & 47.2  & 50.0         & 61.1        & 74.1   & 14.5  & 30.5  & 74.2  & 89.0    & 103.7       & 120.6       & 140.9\\
    LDR \cite{LDR}                   & 13.1  & 23.7  & 44.5  & 50.9  & \textbf{—}    & \textbf{—}      & 78.3   & 7.6   & 15.9  & 37.2  & 41.7  & \textbf{—}  & \textbf{—} & 53.8   & 9.6   & 27.9  & 66.3  & 78.8    & \textbf{—} & \textbf{—} & 129.7\\
    TrajNet \cite{TrajNet}           & 9.7   & 22.3  & 50.2  & 61.7  & 84.7           & \textbf{—}      & 104.2  & 8.5   & 18.4  & 37.0  & 44.8  & 59.2         & \textbf{—} & 71.5   &12.6   & 28.1  & 67.3  & 80.1    & 91.4        & \textbf{—} & 84.3 \\
    SDMTL\cite{SDMTL}                & 9.8   & 23.4  & 53.8  & 67.0  & 88.3           & \textbf{—}      & 107.9  & 8.2   & 16.4  & 33.8  & 42.4  & 53.9         & \textbf{—} & 68.8   & 11.7  & 25.3  & 61.9  & 75.0    & 88.7        & \textbf{—} & 89.0   \\
    HRI \cite{HRI}                   & 7.4   & 18.4  & 44.5  & 56.5  & 73.9           & 88.2             & 106.5  & 6.4   & 14.0  & 28.7  & 36.2  & 50.0         & 61.4        & 75.7   & 13.7  & 30.1  & 63.8  & 78.1    & 101.9       & 118.4       & 138.8\\

    \midrule
    Our  & \textbf{5.9} & \textbf{14.2} & \textbf{37.6} & \textbf{42.5} & \textbf{64.8} & \textbf{71.6} & \textbf{72.3} & \textbf{4.7} & \textbf{10.8} & \textbf{21.0} & \textbf{28.2} & \textbf{36.3} & \textbf{43.9} & \textbf{52.5} & \textbf{7.9} & \textbf{18.4} & \textbf{46.8} & \textbf{55.2} & \textbf{68.2} & \textbf{75.8} & \textbf{83.1}  \\\hline
    \hline
    \multicolumn{2}{r}{} &     & \multicolumn{4}{c}{Sitting} & \multicolumn{2}{c}{} &  & \multicolumn{4}{c}{Sitting Down} & \multicolumn{2}{c}{} &       & \multicolumn{4}{c}{Taking Photo}  \\
    Millisecond(ms) & 80    & 160   & 320   & 400    & 560    & 720   & 1,000  & 80    & 160   & 320   & 400  & 560    & 720   & 1,000 & 80    & 160   & 320   & 400  & 560    & 720   & 1,000  \\
    \midrule
    Res-GRU \cite{residual}          & 23.8  & 44.7  & 78.0  & 91.2  & 113.7         & 130.5          & 152.6 & 31.7  & 58.3  & 96.7  & 112.0 & 138.8        & 159.0         & 187.4  & 21.9  & 41.4  & 74.0  & 87.6   & 110.6        & 128.9        & 153.9 \\
    ConSeq2Seq \cite{Conseq2seq}     & 13.5  & 27.0  & 52.0  & 63.1  & 82.4          & 98.8           & 120.7 & 20.7  & 40.6  & 70.4  & 82.7  & 106.5        & 125.1         & 150.3  & 12.7  & 26.0  & 52.1  & 63.6   & 84.4         & 102.4        & 128.1 \\
    HMR \cite{cvprliu}               & 12.6  & 25.6  & 44.7  & 60.7  & 76.4          & 96.3           & 118.4 & 9.6   & 18.6  & 41.1  & 57.7  & 101.7        & 128.8         & 148.3  & 7.9   & 19.0  & 31.5  & 57.3   & 83.5         & 93.7         & 108.5 \\
    FC-GCN \cite{FC-GCN}             & 10.7  & 24.6  & 50.6  & 62.0  & 76.4          & 93.1           & 115.7 & 11.4  & 27.6  & 56.4  & 67.6  & 96.2         & 115.2         & 142.2  & 6.8   & 15.2  & 38.2  & 49.6   & 72.5         & 90.9         & 116.3 \\
    LDR \cite{LDR}                   & 9.2   & 23.1  & 47.2  & 57.7  & \textbf{—}   & \textbf{—}    & 106.5 & 9.3   & 21.4  & 46.3  & 59.3  & \textbf{—}  & \textbf{—}   & 144.6  & 7.1   & 13.8  & 29.6  & 44.2   & \textbf{—}  & \textbf{—}  & 116.4 \\
    TrajNet \cite{TrajNet}           & 9.0   & 22.0  & 49.4  & 62.6  & 81.0          & \textbf{—}    & 116.3 &10.7   & 28.8  & 55.1  & 62.9  & 79.8         & \textbf{—}   & 123.8  & 5.4   & 13.4  & 36.2  & 47.0   & 73.0         & \textbf{—}  & 86.6  \\
    SDMTL\cite{SDMTL}                & 8.7   & 22.2  & 52.2  & 65.5  & 83.9          & \textbf{—}    & 115.5 & 9.3   & 23.8  & 50.6  & 60.9  & 77.7         & \textbf{—}   & 118.9  & 6.0   & 14.0  & 36.1  & 47.0   & 67.1         & \textbf{—}  & 91.1  \\
    HRI \cite{HRI}                   & 9.3   & 20.1  & 44.3  & 56.0  & 76.4          & 93.1           & 115.9 & 14.9  & 30.7  & 59.1  & 72.0  & 97.0         & 116.1         & 143.6  & 8.3   & 18.4  & 40.7  & 51.5   & 72.1         & 90.4         & 115.9 \\
    \midrule
    Our  & \textbf{7.6} & \textbf{17.2} & \textbf{36.9} & \textbf{51.2 } & \textbf{69.5} & \textbf{78.3} & \textbf{93.6}  & \textbf{7.2} & \textbf{16.3} & \textbf{32.7} & \textbf{50.9} & \textbf{62.1} & \textbf{94.5} & \textbf{101.9} & \textbf{5.2} & \textbf{12.2} & \textbf{23.8} & \textbf{38.4} & \textbf{62.4} & \textbf{76.2} & \textbf{82.4}\\
    \hline
    \hline
    \multicolumn{2}{r}{} &            & \multicolumn{4}{c}{Phoning} & \multicolumn{2}{c}{} &       & \multicolumn{4}{c}{Posing} & & \multicolumn{2}{c}{}   & \multicolumn{4}{c}{Purchases} \\
    Millisecond(ms) & 80    & 160   & 320   & 400    & 560    & 720   & 1,000  & 80    & 160   & 320   & 400  & 560    & 720   & 1,000 & 80    & 160   & 320   & 400  & 560    & 720   & 1,000  \\
    \midrule
    Res-GRU \cite{residual}       & 21.1  & 38.9  & 66.0  & 76.4  & 94.0          & 107.7        & 126.4          & 29.3  & 56.1  & 98.3   & 114.3  & 140.3         & 159.8         & 183.2    & 28.7  & 52.4  & 86.9   & 100.7 & 122.1        & 137.2        & 154.0 \\
    ConSeq2Seq \cite{Conseq2seq}  & 13.5  & 26.6  & 49.9  & 59.9  & 77.1          & 92.1         & 114.0          & 16.9  & 36.7  & 75.7   & 92.9   & 122.5         & 148.8         & 187.4    & 20.3  & 41.8  & 76.5   & 89.9  & 111.3        & 129.1        & 151.5  \\  
    HMR \cite{cvprliu}            & 12.5  & 21.3  & 39.3  & 58.6  & 71.3          & 88.7         & 112.8          & 13.6  & 23.5  & 62.5   & 114.1  & 126.3         & 135.9         & 143.6    & 15.3  & 30.6  & 64.7   & 73.9  & 97.5         & 107.2        & 122.7   \\
    FC-GCN \cite{FC-GCN}          & 11.5  & 20.2  & 37.9  & 43.2  & 67.8          & 83.0         & 105.1          & 9.4   & 23.9  & 66.2   & 82.9   & 107.6         & 136.1         & 175.0    & 19.6  & 38.5  & 64.4   & 72.2  & 98.3         & 115.1        & 139.3   \\
    LDR \cite{LDR}                & 10.4  & 14.3  & 33.1  & 39.7  & \textbf{—}   & \textbf{—}  & \textbf{85.8}  & 8.7   & 21.1  & 58.3   & 81.9   & \textbf{—}   & \textbf{—}   & 133.7    & 16.2  & 36.1  & 62.8   & 76.2  & \textbf{—}  & \textbf{—}  & 112.6 \\
    TrajNet \cite{TrajNet}        & 10.7  & 18.8  & 37.0  & 43.1  & 62.3          & \textbf{—}  & 113.5          & 6.9   & 21.3  & 62.9   & 78.8   & 111.6         & \textbf{—}   & 210.9    & 17.1  & 36.1  & 64.3   & 75.1  & 84.5         & \textbf{—}  & 115.5\\
    SDMTL \cite{SDMTL}            & 10.5  & 18.5  & 37.2  & 43.1  & 60.8          & \textbf{—}  & 112.3          & 6.8   & 20.5  & 64.0   & 82.4   & 107.2         & \textbf{—}   & 204.7    & 18.4  & 38.8  & 61.1   & 68.2  & 80.9         & \textbf{—}  & 113.6\\
    HRI \cite{HRI}                & 8.6   & 18.3  & 39.0  & 49.2  & 67.4          & 82.9         & 105.0          & 10.2  & 24.2  & 58.5   & 75.8   & 107.6         & 136.8         & 178.2    & 13.0  & 29.2  & 60.4   & 73.9  & 95.6         & 110.9        & 134.2  \\
    \midrule
    Our    & \textbf{6.6} & \textbf{10.1} & \textbf{24.3} & \textbf{31.4} & \textbf{48.2} & \textbf{74.8} & 102.7 & \textbf{5.2} & \textbf{16.9} & \textbf{49.2} & \textbf{68.3} & \textbf{96.6} & \textbf{118.0} & \textbf{123.8} & \textbf{10.2} & \textbf{23.6} & \textbf{52.3} & \textbf{58.6} & \textbf{73.0} & \textbf{92.7} & \textbf{112.2} \\
   \hline
    \hline
    \multicolumn{2}{r}{} &       & \multicolumn{4}{c}{Waiting} & \multicolumn{2}{c}{} &       & \multicolumn{4}{c}{Walking Dog} &
    \multicolumn{2}{c}{} &       & \multicolumn{4}{c}{Average} &  \\
    Millisecond(ms) & 80    & 160   & 320   & 400    & 560    & 720   & 1,000  & 80    & 160   & 320   & 400  & 560    & 720   & 1,000 & 80    & 160   & 320   & 400  & 560    & 720   & 1,000  \\
    \midrule
    Res-GRU \cite{residual}      & 23.8  & 44.2  & 75.8  & 87.7  & 105.4        & 117.3        & 135.4  & 36.4  & 64.8  & 99.1   & 110.6  & 128.7       & 141.1         & 164.5  & 25.0   & 46.2    & 77.0  & 88.3  & 106.3       & 119.4         & 136.6\\
    ConSeq2Seq \cite{Conseq2seq} & 14.6  & 29.7  & 58.1  & 69.7  & 87.3         & 100.3        & 117.7  & 27.7  & 53,6  & 90.7   & 103.3  & 122.4       & 133.8         & 162.4  & 16.6   & 33.3    & 61.4  & 72.7  & 90.7        & 104.7         & 124.2  \\
    HMR \cite{cvprliu}           & 12.8  & 24.5  & 45.2  & 85.1  & 87.5         & 94.2         & 121.9  & 30.1  & 41.4  & 78.4   & 100.1  & 134.7       & 141.6         & 157.4  & 13.3   & 23.2    & 44.7  & 63.8  & 86.1        & 99.9          & 116.2  \\
    FC-GCN \cite{FC-GCN}         & 9.5   & 22.0  & 57.5  & 73.9  & 73.4         & 88.2         & 107.5  & 32.2  & 58.0  & 102.2  & 122.7  & 105.8       & 118.7         & 142.2  & 12.1   & 25.0    & 51.0  & 61.3  & 78.3        & 93.3          & 114.0 \\
    LDR \cite{LDR}               & 9.2   & 17.6  & 47.2  & 71.6  & \textbf{—}  & \textbf{—}  & 127.3  & 25.3  & 56.6  & 87.9   & 99.4   & \textbf{—} & \textbf{—}   & 143.2  & 10.7   & 22.5    & 45.1  & 55.8  & \textbf{—} & \textbf{—}  & 97.8\\
    TrajNet \cite{TrajNet}       & 8.2   & 21.0  & 53.4  & 68.9  & 92.9         & \textbf{—}  & 165.9  & 23.6  & 52.0  & 98.1   & 116.9  & 141.1       & \textbf{—}   & 181.3  & 10.2   & 23.2    & 49.3  & 59.7  & 77.7        & \textbf{—}   & 110.6  \\
    SDMTL \cite{SDMTL}           & 7.5   & 19.0  & 46.8  & 58.3  & 81.4         & \textbf{—}  & 159.2  & 21.0  & 54.9  & 100.4  & 119.8  & 137.7       & \textbf{—}   & 181.5  & 9.8    & 22.7    & 48.0  & 58.2  & 74.5        & \textbf{—}   & 110.7  \\
    HRI \cite{HRI}               & 8.7   & 19.2  & 43.4  & 54.9  & 74.5         & 89.0         & 108.2  & 20.1  & 40.3  & 73.3   & 86.3   & 108.2       & 120.6         & 146.9  & 10.4   & 22.6    & 47.1  & 58.3  & 77.3        & 91.8          & 112.1\\
    \midrule
    Our   & \textbf{6.5} & \textbf{15.2} & \textbf{37.5} & \textbf{47.3} & \textbf{68.8} & \textbf{79.4} & \textbf{95.6} & \textbf{15.5} & \textbf{31.9} & \textbf{62.3} & \textbf{67.4} & \textbf{85.6} & \textbf{106.3} & \textbf{126.1} & \textbf{8.6} & \textbf{19.6} & \textbf{39.2} & \textbf{50.5} & \textbf{68.9} & \textbf{81.3} & \textbf{93.8}  \\
   \hline
    \end{tabular}%
  }

\label{table1}
\end{table*}

\begin{figure}[t]
\begin{center}
\includegraphics[width=0.9\linewidth]{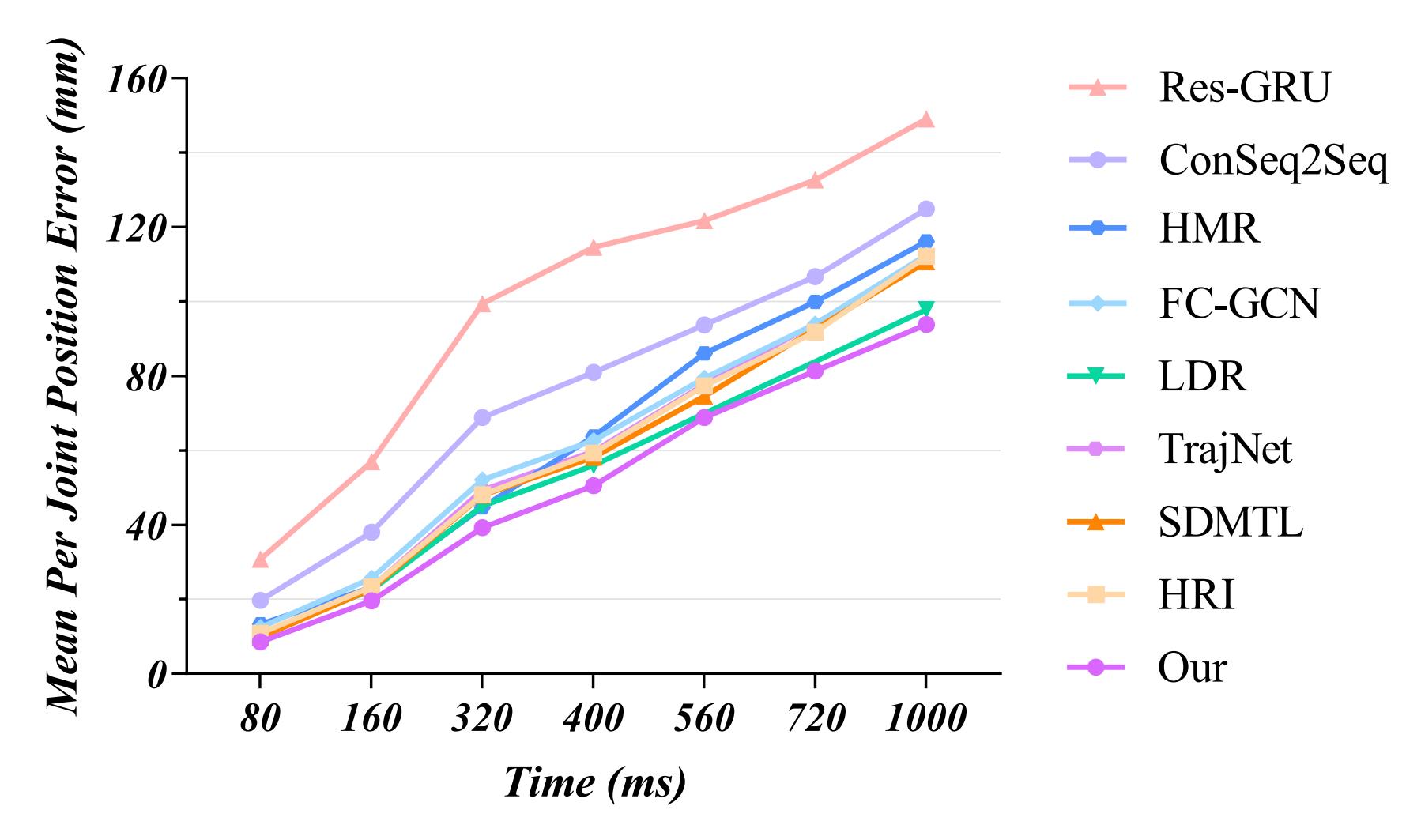}
\end{center}
   \caption{The average Mean Per Joint Position Error for all 15 actions on H3.6m dataset.}
\label{average}
\end{figure}

\begin{figure*}[htbp]
\begin{center}
\includegraphics[width=1\textwidth]{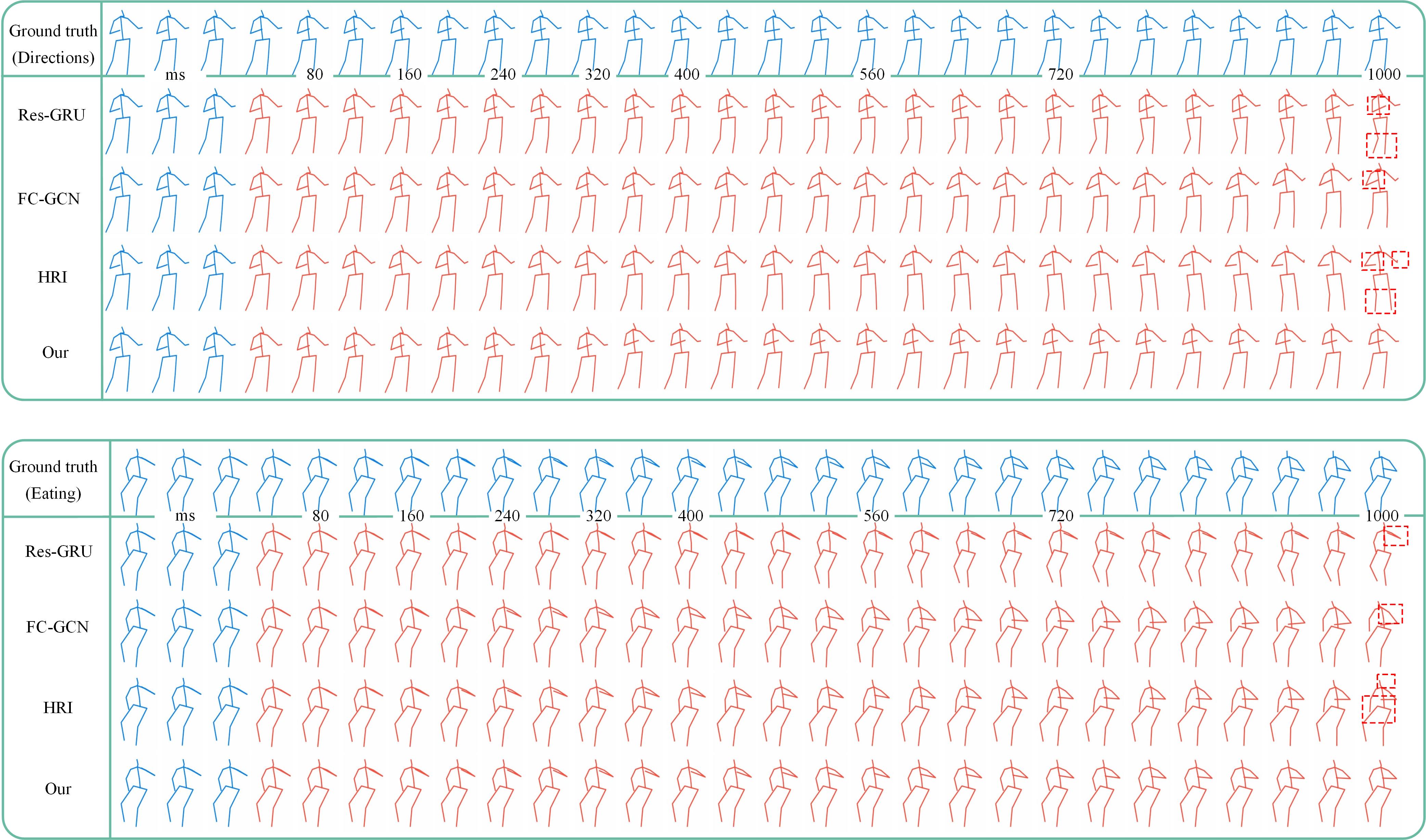}
\end{center}
\caption{The visual results comparison on H3.6M. In each sub-group, the first row shows the ground truth, and the following rows are the results of Res-GRU, FC-GCN, HRI, and our method.}
\label{vis_results}
\end{figure*}

\subsection{Evaluation on Human Datasets}
We evaluate our method and existing works on two popular benchmark datasets. For fair comparison, we follow the evaluation metric in existing works \cite{FC-GCN, HRI}. Specifically, the standard Mean Per Joint Position Error (MPJPE) is adopted to measure the mean Euclidean distance between the predicted joint positions and the ground truth. The results are reported on both short-term ($80-400$ms) and long-term ($400-1,000$ms).


\textbf{H3.6m} \quad We benchmark the proposed method against 8 existing works, including Res-GRU \cite{residual}, ConSeq2Seq \cite{Conseq2seq}, HMR \cite{cvprliu}, FC-GCN \cite{FC-GCN}, TrajNet \cite{TrajNet}, SDMTL\cite{SDMTL}, LDR \cite{LDR}, and HRI \cite{HRI}. The experimental results are demonstrated in Table~\ref{table1} and Fig.~\ref{average}. We report the results of 11 diverse actions and the average over all actions.  There are different levels of complexity for different activities. Existing methods tend to perform well for motions with high degrees of periodicity such as ``Walking" or regularity such as ``Eating". However, their performance suffers major dips when modeling more stochastic and irregular movements such as ``Posing" or ``Directions". On the contrary, a major highlight of our model is the ability to deliver   accurate predictions even when faced with highly complex and aperiodic action types. We observe that our method consistently delivers superior performance over state-of-the-art methods, which achieves a substantial 15.4\% accuracy improvement in average comparing to state-of-the-art approach. The improvement is even more significant  for long-term predictions. 

We further visualize sample motion prediction results in Fig.~\ref{vis_results}, where the results for ``Directions" and ``Eating" activities are presented. The first three frames correspond to the observed sequences and the subsequent frames are the predicted poses. Each row illustrates the results of a method. It is easy to see that the forecasted sequences obtained by our method are more similar with the ground truth as well as being more natural. A significant issue arising in existing methods for highly stochastic actions is that the predicted motions tend to converge to static motionless states. For example, the HRI \cite{HRI} predictions for ``Directions" have very limited range of motions, which appears unnatural. Incorporating displacements and adopting a joint phase space trajectory representation alleviates this plaguing issue, as it provides richer and smoother individualized motion contexts that are easier to model than the pose representation. In the illustrations of the ``Eating" action, the hands and foot coordination for our method also appears more coherent. We attribute this to our explicit modeling of the joint dependencies that serves to avoid interference and excess of information from the closely correlated joints.

\textbf{CMU MoCap} \quad We also evaluate our method on the CMU MoCap dataset. As shown in Table~\ref{table2}, our method consistently and substantially outperforms all the  existing methods, especially for complex activity types. For example, on complex activity types such as ``Basketball", ``Directing Traffic",  and ``Soccer", our method significantly outperforms state-of-the-art methods by 17.4\%, 14.7\%, and 25.4\% , respectively. 

\begin{figure*}[htbp]
\begin{center}
\includegraphics[width=1\textwidth]{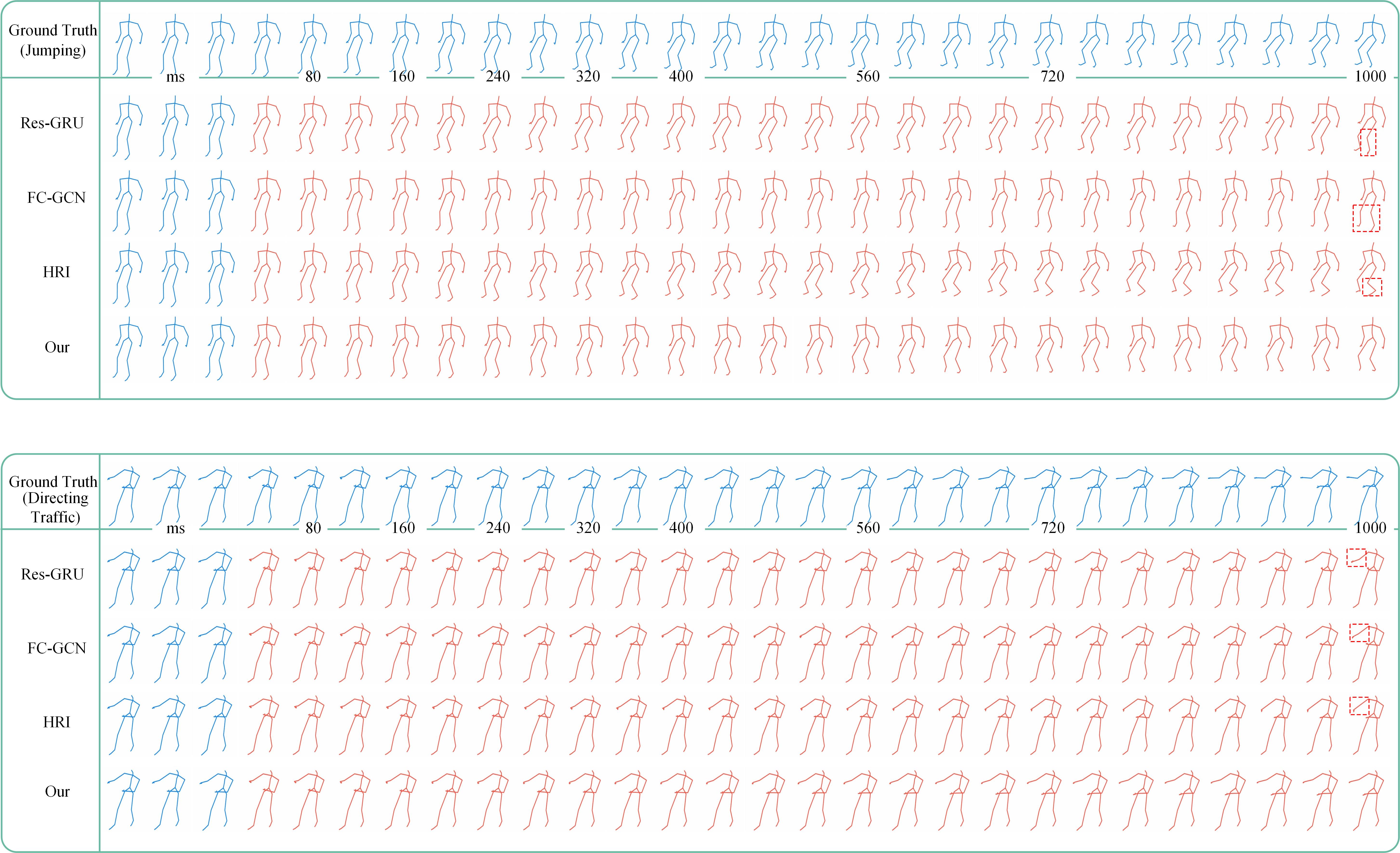}
\end{center}
\caption{The visual results comparison on CMU-MoCap dataset. In each sub-group, the first row shows the ground truth, and the following rows are the results of Res-GRU, FC-GCN, HRI, and our method.}
\label{cmu_vis_results}
\end{figure*}

For further evaluation, we show the visual results on the CMU-MoCap dataset in Fig.~\ref{cmu_vis_results}. Our method outperforms all state-of-the-art methods in both long-term and short-term predictions and is proven to be effective and robust in capturing temporal dynamics. For the ``Jumping" action, our model successfully captures the fast rise of legs while other methods fail to extract such a trend. More specifically, Res-GRU and FC-GCN fail to predict the subtle correlation between the legs; HRI generates unreasonable prediction for the left foot. For the ``Directing Traffic" action, we can observe that our method generates more accurate hand motions than state-of-the-art methods, while Res-GRU exhibits discontinuity between the first frame of prediction and the historical sequence; HRI yields good short-term predictions but is not accurate in the long-term.


\begin{table*}[t]
\caption{Comparisons of position error for both short-term and long-term predictions on CMU-MoCap dataset with state-of-the-art methods.}
  \centering
  \resizebox{0.97\textwidth}{!}{
    \begin{tabular}{c|cccccc|cccccc|cccccc|cccccc}
    \hline
        \multicolumn{2}{r}{} &       & \multicolumn{3}{c}{Basketball} & \multicolumn{1}{c}{} &       & \multicolumn{4}{c}{Basketball signal} & \multicolumn{2}{c}{} &       & \multicolumn{3}{c}{Directing traffic} &
    \multicolumn{2}{c}{} &       & \multicolumn{3}{c}{Jumping} &
    \\
    Millisecond(ms) & 80    & 160   & 320   & 400 & 560  & 1,000  & 80    & 160   & 320   & 400  & 560  & 1,000 & 80    & 160   & 320   & 400  & 560  & 1,000  & 80    & 160   & 320   & 400  & 560  & 1,000 \\
    \midrule
    Res-GRU \cite{residual}  & 18.4  & 33.8  & 59.5  & 70.5  & —     & 106.7  & 12.7  & 23.8  & 40.3  & 46.7  & \textbf{—}     & 77.5  & 15.2  & 29.6  & 55.1  & 66.1  & \textbf{—}     & 127.1 & 36.0  & 68.7  & 125.0  & 145.5  & \textbf{—}     & 195.5  \\
    ConSeq2Seq \cite{Conseq2seq}& 16.7  & 30.5  & 53.8  & 64.3  & —     & 91.5  & 8.4   & 16.2  & 30.8  & 37.8  & \textbf{—}     & 76.5  & 10.6  & 20.3  & 38.7  & 48.4  & \textbf{—}     & 115.5 & 22.4  & 44.0  & 87.5  & 106.3  & \textbf{—}     & 162.6  \\
    FC-GCN \cite{FC-GCN} & 14.0  & 25.4  & 49.6  & 61.4  & 77.4  & 106.1  & 3.5   & 6.1   & 11.7  & 15.2  & 25.3  & 53.9  & 7.4   & 15.1  & 31.7  & 42.2  & 70.3  & 152.4 & 16.9  & 34.4  & 76.3  & 96.8  & 131.4  & 164.6  \\
    LPJP \cite{transformer}  & 11.6  & 21.7  & 44.4  & 57.3   & \textbf{—}  & 90.9  & 2.6  & 4.9   & 12.7   &  18.7  & \textbf{—}  & 75.8   & 6.2   &  12.7   & 29.1   &  39.6  & \textbf{—}  & 149.1  & 12.9 & 27.6 & 73.5 & 92.2 & \textbf{—}  &  176.6  \\
    LDR \cite{LDR} & 13.1  & 22.0  & 37.2  & 55.8  & \textbf{—}     & 97.7  & 3.4   & 6.2   & 11.2  & 13.8  & \textbf{—}     & 47.3  & 6.8   & 16.3  & 27.9  & 38.9  & \textbf{—}     & 131.8 & 13.2  & 32.7  & 65.1  & 91.3  & \textbf{—}     & 153.5  \\
    SDMTL \cite{SDMTL}  & 10.9  & 20.2  & 40.9  & 50.8  & 66.1  & 110.2  & 2.9   & 6.2   & 16.4  & 23.1  & 37.4  & 71.6  & 5.1   & 10.9  & 23.2  & 30.2  & 46.1  & 104.5 & 11.1  & 24.6  & 65.7  & 90.3  & 130.9  & 191.2  \\
    \midrule
    Our    & \textbf{10.7} & \textbf{16.9} & \textbf{33.4} & \textbf{42.9} & \textbf{57.3} & \textbf{88.5} & \textbf{2.1} & \textbf{4.9} & \textbf{9.2} & \textbf{10.4} & \textbf{19.8} & \textbf{44.3} & \textbf{4.3} & \textbf{9.5} & \textbf{18.9} & \textbf{27.6} & \textbf{40.0} & \textbf{92.4} & \textbf{10.6} & \textbf{20.4} & \textbf{52.0} & \textbf{80.8} & \textbf{114.9} & \textbf{146.0} \\ \hline
    \hline
    \multicolumn{2}{c}{} &       & \multicolumn{3}{c}{Soccer} & \multicolumn{1}{c}{} &       & \multicolumn{4}{c}{Walking} & \multicolumn{1}{c}{} &       & \multicolumn{4}{c}{Wash window} &
    \multicolumn{2}{c}{} &       & \multicolumn{2}{c}{Average} &
     \\
    Millisecond(ms) & 80    & 160   & 320   & 400  & 560  & 1,000  & 80    & 160   & 320   & 400  & 560   & 1,000 & 80    & 160   & 320   & 400  & 560   & 1,000  & 80    & 160   & 320   & 400  & 560   & 1,000 \\
    \midrule
    Res-GRU \cite{residual} & 20.3  & 39.5  & 71.3  & 84.0  & \textbf{—}     & 129.6  & 8.2   & 13.7  & 21.9  & 24.5  & \textbf{—}     & 52.2  & 8.4   & 15.8  & 29.3  & 35.4  & \textbf{—}     & 61.1  & 16.8   & 30.5   & 54.2   & 63.6   & \textbf{—}    & 99.0 \\
    ConSeq2Seq \cite{Conseq2seq} & 12.1  & 21.8  & 41.9  & 52.9  & \textbf{—}     & 94.6  & 7.6   & 12.5  & 23.0  & 27.5  & \textbf{—}     & 49.8  & 8.2   & 15.9  & 32.1  & 39.9  & \textbf{—}    & 58.9  &  12.5 & 22.2   & 40.7  & 49.7  & \textbf{—}  & 84.6 \\
    FC-GCN \cite{FC-GCN}& 11.3  & 21.5  & 44.2  & 55.8  & 82.6  & 117.5  & 7.7   & 11.8  & 19.4  & 23.1  & 27.2  & 40.2  & 5.9   & 11.9  & 30.3  & 40.0  & 53.0  & 79.3  & 11.5  & 20.4  & 37.8  & 46.8  & 62.9  & 96.5 \\
    LPJP \cite{transformer}  & 9.2  & 18.4  & 39.2  & 49.5  & \textbf{—}  & 93.9  & 6.7  & 10.7   & 21.7   & 27.5  & \textbf{—}  & \textbf{37.4} & 5.4   & 11.3    & 29.2   & 39.6   & \textbf{—}  & 79.1  & 9.8  & 17.6  & 35.7  & 45.1  & \textbf{—}  & 93.2  \\
    LDR \cite{LDR}  & 10.3  & 21.1  & 42.7  & 50.9  & \textbf{—}     & 91.4  & 7.1   & 10.4  & 17.8  & 20.7  & \textbf{—}     & 37.5  & 5.8   & 12.3  & 27.8  & 38.2  & \textbf{—}     & 56.6  & 9.4   & 18.8  & 31.6  & 43.2  & \textbf{—}     & 82.9 \\
    SDMTL \cite{SDMTL} & 8.1   & 16.5  & 36.6  & 50.6  & 77.0  & 140.7  & 6.1   & 9.0   & 17.5  & 20.0  & 26.3  & 51.9  & 4.6   & 10.1  & 29.6  & 39.2  & 50.9  & 82.4  & 8.0   & 14.5  & 31.9  & 41.9  & 59.4  & 102.7 \\
    \midrule
    Our & \textbf{6.5} & \textbf{12.5} & \textbf{26.3} & \textbf{40.6} & \textbf{68.1} & \textbf{81.5} & \textbf{5.3} & \textbf{7.8} & \textbf{15.9} & \textbf{18.0} & \textbf{25.5} & 44.7 & \textbf{4.2} & \textbf{8.5} & \textbf{24.3} & \textbf{32.6} & \textbf{46.0} & \textbf{55.4} & \textbf{6.6}     & \textbf{12.4}  & \textbf{27.0}      & \textbf{36.6}      & \textbf{51.4}      & \textbf{76.2} \\
    \bottomrule
    \end{tabular}
  \label{tab:addlabel}
}

\label{table2}
\end{table*}

\subsection{Ablation Studies}
We investigate the effectiveness of different modules within our TEID through the following ablation studies. Experiments are performed on the H3.6m dataset, with results reported in Table~\ref{table4}.

{\bf Explicit dependency modeling module} \quad The explicit dependency modeling block in our TEID serves to explicitly account for prior knowledge such as natural anatomical connection between joints and coordination of different limbs. From the results in Table~\ref{table4}, we observe that the removal of this module will bring slight accuracy decay in the short-term while the long-term prediction accuracy is greatly affected. We interpret this as the failure to leverage prior anatomy knowledge will be detrimental in effectively filtering out joint correlations, resulting in error accumulation for the long-term. 

{\bf Implicit global optimization module} \quad This module is designed to globally optimize the position of a joint with respect to the entire forecasted motion sequence. Removing this block results in slight dip in accuracy. This reveals that modeling only explicitly related joints while ignoring irrelevant joints is able to deliver relatively accurate motion prediction. However, implicit relation modeling considers the hidden dependencies between each pair of joints ( \emph{e.g.}, the potential correlation between left hand and right hand joints), which improves the naturalness of the predicted pose sequence.


{\bf Phase space trajectory representation} \quad Instead of explicitly incorporating the displacement vector as inputs, we also try employing only the position vectors for each joint. From the quantitative results in Table~\ref{table4}, we observe a significant deterioration in prediction accuracy, testifying to the empirical effectiveness of including displacements. This agrees with our intuition that a complete characterization of the dynamic system with a phase space representation plays an important role in motion prediction.

\begin{table}[t]
\caption{Ablation studies for the \textbf{E}xplicit denpendency modeling module, \textbf{I}mplicit dependency modeling module, and \textbf{D}isplacement inputs. E and I stand for Explicit and Implicit dependency modeling modules, respectively. D represents displacement inputs.}
\centering
\resizebox{1\columnwidth}{!}{
\begin{tabular}{ccc|ccccccc}
\toprule
\multicolumn{1}{c}{\textbf{E}}&\multicolumn{1}{c}{\textbf{I}} &\multicolumn{1}{c|}{\textbf{D}} & 80 & 160 & 320 & 400 & 560 & 720 & 1,000 \\
\midrule
\multicolumn{1}{c}{\textbf{\checkmark}} & \multicolumn{1}{c}{\textbf{\checkmark}} & \multicolumn{1}{c|}{} & 28.9 & 34.8 & 52.6 & 65.3 & 88.1 & 102.0 & 117.5 \\
\multicolumn{1}{c}{} & \multicolumn{1}{c}{\textbf{\checkmark}}& \multicolumn{1}{c|}{\textbf{\checkmark}} & 9.5 & 24.5 & 44.7 & 60.3 & 82.8 & 93.6 & 105.3 \\
\multicolumn{1}{c}{\textbf{\checkmark}} & \multicolumn{1}{c}{} & \multicolumn{1}{c|}{\textbf{\checkmark}} & 9.2 & 23.3 & 42.5 & 54.7 & 75.1 & 86.0 & 95.2 \\
\multicolumn{1}{c}{\textbf{\checkmark}} & \multicolumn{1}{c}{\textbf{\checkmark}}& \multicolumn{1}{c|}{\textbf{\checkmark}} & \textbf{8.6} & \textbf{19.6} & \textbf{39.2} & \textbf{50.5} & \textbf{68.9} & \textbf{81.3} & \textbf{93.8} \\
\hline
\end{tabular}
}

\label{table4}
\end{table}

\section{Conclusion}
In this paper, we tackle the motion prediction problem by moving away from kinematics graph pose representations and instead adopting a phase space trajectory representation for each constituent joint. This serves to reduce the inherent complexity of the problem by considering individualized joint trajectories instead of the entire pose sequence. We further design a network consisting of prior anatomical knowledge encoding and multi-scale convolution  for explicit joint dependency modeling. Along with a global affinity-based optimization module, we obtain joint trajectory extrapolations that aggregate coherently to form a consistent and natural pose sequence. Our method is robust and accurate, demonstrating significant improvements over state-of-the-art methods.

\section{Acknowledgments} 
This research is supported in part by the National Key Research and Development Program of China under Grant  No.2020AAA0140004, the Natural Science Foundation of Zhejiang Province, China (Grant No. LQ19F020001), the National Natural Science Foundation of China (No. 61902348), and the Key R\&D Program of Zhejiang Province (No. 2021C01104).

%




\bibliographystyle{ACM-Reference-Format}
\balance
\bibliography{egbib}

\end{document}